\documentclass[lettersize,journal]{IEEEtran}
\usepackage{amsmath,amsfonts}
\usepackage{algorithmic}
\usepackage{array}
\usepackage[caption=false,font=normalsize,labelfont=sf,textfont=sf]{subfig}
\usepackage{textcomp}
\usepackage{stfloats}
\usepackage{url}
\usepackage{verbatim}
\usepackage{graphicx}
\usepackage{cite}
\usepackage{multirow}
\usepackage[ruled,linesnumbered,boxed]{algorithm2e}  
\usepackage[numbers]{natbib}
\usepackage{ulem}
\usepackage{booktabs}
\usepackage{bm}
\usepackage{enumerate}
\begin{document}

\title{Spectrum-BERT: Pre-training of Deep Bidirectional Transformers for Spectral Classification of Chinese Liquors}

\author{$\text{Yansong Wang}^\dag$, $\text{Yundong Sun}^\dag$, Yansheng Fu, $\text{Dongjie Zhu}^\ast$, Zhaoshuo Tian,~\IEEEmembership{Member,~IEEE}

\thanks{\dag Equal contribution.}
\thanks{This work is supported by the National Key R\&D Program of China (2020YFE0201500), the Fundamental Research Funds for the Central Universities (HIT.NSRIF.201714). (\emph{Corresponding author: Dongjie Zhu.})}
\thanks{Yansong Wang, Yansheng Fu and Dongjie Zhu are with the School of Computer Science and Technology at Harbin Institute of Technology at Weihai, Weihai 264209, China (e-mail: yansongwang0629@163.com; fuyansheng123@gmail.com; zhudongjie@hit.edu.cn.)}
\thanks{Yundong Sun is with the Department of Electronic Science and Technology at Harbin Institute of Technology, Harbin 150001, China, and also with the School of Computer Science and Technology at Harbin Institute of Technology at Weihai, Weihai 264209, China (e-mail: hitffmy@163.com).}
\thanks{Zhaoshuo Tian is with the Institute of Marine Optoelectronic Equipment, Harbin Institute of Technology at Weihai, Weihai, 264209, China (e-mail: tianzhaoshuo@126.com).}

}

\markboth{IEEE Transactions on Instrumentation and Measurement}%
{Shell \MakeLowercase{\textit{et al.}}: A Sample Article Using IEEEtran.cls for IEEE Journals}


\maketitle

\begin{abstract}
The frequent occurrence of Chinese liquor counterfeiting incidents has disturbed market operation orders and even seriously damaged consumers' health. Spectral detection technology, as a non-invasive method for rapid detection of substances, combined with deep learning algorithms, has been widely used in food detection. However, in real scenarios, acquiring and labeling spectral data is an extremely labor-intensive task, which makes it impossible to provide enough high-quality data for training efficient supervised deep learning models. To better leverage limited samples, we apply "\normalem{\emph{pre-training \& fine-tuning}}" paradigm to the field of spectral detection for the first time and propose a pre-training method of deep bidirectional transformers for spectral classification of Chinese liquors, abbreviated as Spectrum-BERT. Specifically, first, to retain the model's sensitivity to the characteristic peak position and local information of the spectral curve, we innovatively partition the curve into multiple blocks and obtain the embeddings of different blocks, as the feature input for the next calculation. Second, in the pre-training stage, we elaborately design two pre-training tasks, Next Curve Prediction (NCP) and Masked Curve Model (MCM), so that the model can effectively utilize unlabeled samples to capture the potential knowledge of spectral data, breaking the restrictions of the insufficient labeled samples, and improving the applicability and performance of the model in practical scenarios. Finally, we conduct a large number of experiments on the real liquor spectral dataset. In the comparative experiments, the proposed Spectrum-BERT significantly outperforms the baselines in multiple metrics and this advantage is more significant on the imbalanced dataset. Moreover, in the parameter sensitivity experiment, we also analyze the model performance under different parameter settings, to provide a reference for subsequent research.
 
\end{abstract}

\begin{IEEEkeywords}
Spectral detection, liquor detection, pre-training, few-shot learning, BERT, deep learning.
\end{IEEEkeywords}

\section{Introduction}
\IEEEPARstart{A}{t} present, food safety incidents occur frequently all over the world, which has seriously damaged the normal market operation order, violated the rights and interests of regular manufacturers and consumers, and even caused serious damage to the health of consumers \cite{tian2022development}. According to the World Health Organization, unsafe food can cause 600 million illnesses and 50,000 deaths worldwide each year\footnote{https://www.who.int/westernpacific/health-topics/food-safety}. Because of the low cost and high profit of Chinese liquors counterfeiting, it has become the most serious food safety problem.
	
\begin{figure}
	\centering
	\includegraphics[width=2.5in,trim=0 20 0 0]{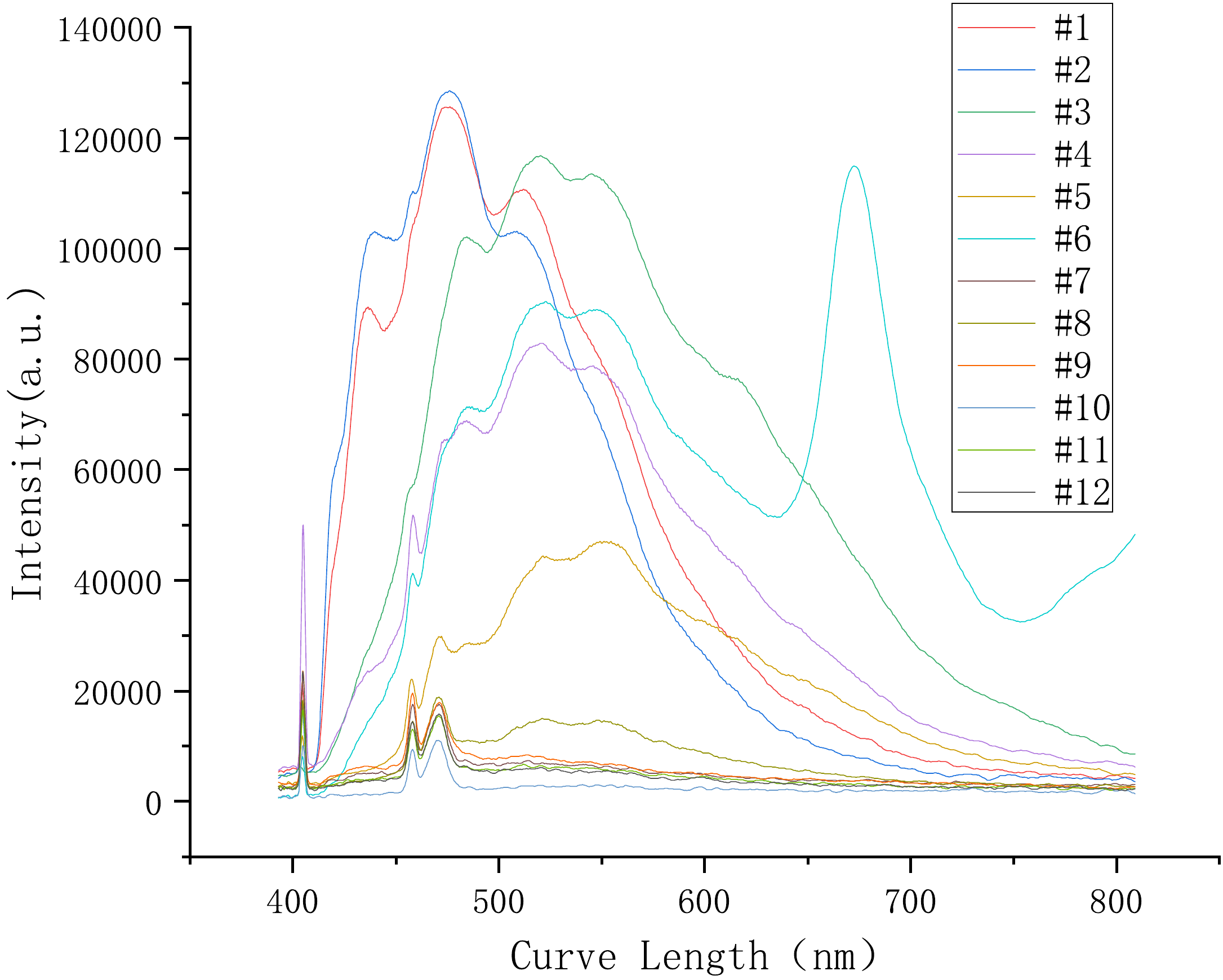}
	\caption{The examples of spectral curves from different Chinese liquors.}
	\label{FIG:1}
		\vspace{-1em}
\end{figure}

Traditional detection methods, such as chromatography \cite{sun2003application,wang2020recent} and electronic nose \cite{hou2021hand,shi2021lightweight}, are relatively demanding on the environment, and are easily affected by the temperature and chemical substances of the detected objects, resulting in inaccurate results. Meanwhile, the detection process is complicated, and the cost of a single experiment is high. Spectral detection technology is widely used in food detection due to its advantages of simple operation, high sensitivity, long life, low cost, and no pollution to the environment and samples \cite{cozzolino2022advantages,yin2022wide}. Fig.\ref{FIG:1} shows the spectral curves of different liquors. It can be clearly seen that the distribution of the obtained spectral curves is different due to the different substance types and contents of different liquors. Therefore, algorithms based on spectral curves can be used for liquor authenticity detection.

Traditional matching algorithms \cite{jin2016use,qin2010raman,qin2017subsurface}, based on spectral curve characteristic peak matching, dominate the existing food spectrum detection. However, those methods require manual extraction, setting, and adjustment of characteristic peaks based on domain knowledge, without the ability to learn and extract features independently. At the same time, the detection target must be compared with a large number of samples in the database, which is also time-consuming and cannot be applied in massive data scenarios. The deep learning algorithm can realize complex feature extraction and predictive analysis tasks without manual feature extraction, and it has better generalization ability than artificially designed features. In recent years, the combination of spectral detection technology and deep learning algorithms has significantly improved the performance of food detection.

However, the existing deep learning spectrum detection models generally fall into the category of supervised learning, requiring enormous labeled datasets for training. The quantity and quality of training samples directly determine the performance of the model. On the one hand, in the field of Chinese liquor detection, there is currently no public liquor spectral dataset, and it is impossible to leverage the existing large-scale general datasets for model training as in natural language processing (NLP) and computer vision (CV). On the other hand, there are considerable brands and types of existing liquors on the market, and the price is high. It is difficult for the researchers to build and label the spectrum dataset of liquors, and the quality of the samples cannot be guaranteed. In conclusion, there is not enough high-quality data in practical application scenarios to support training efficient deep learning models, i.e. few-shot learning problems. As the most effective scheme for few-shot learning, the "\normalem{\emph{pre-training \& fine-tuning}}" paradigm has achieved great success in NLP and CV \cite{2021Knowledge,carion2020end,chen2020generative,he2022masked,zhu2021deformable,liu2020k,wang2021kepler}, but no researchers have tried it in spectral detection. Therefore, it is urgent to propose a new model for spectral detection, which can enable researchers to make full use of unlabeled data to conduct unsupervised pre-training, so that the model can capture the latent knowledge of spectral data in advance, providing an efficient base model for subsequent downstream tasks such as curve classification.

Convolutional Neural Network (CNN) has recently become a hot spot for spectral curve classification, and many researchers \cite{liu2021efficient,yan2021nondestructive,zhu2020deep} have improved the accuracy of spectral detection based on CNN. However, the previous study\cite{liu2017deep} has also shown that the Max-Pooling operation used in CNN may cause the model to lose local information in the process of downsampling. The spatial translation invariance of CNN also makes the captured spectral curve features lose the sensitivity to the position of characteristic peaks. Therefore, we need a new curve feature extraction approach that can accurately capture the local features of the spectral while maintaining good sensitivity to the position of the characteristic peaks.

To solve the aforementioned problems, in this paper, we propose the \textbf{Spectrum-BERT} model inspired by the pre-training idea of the BERT model \cite{BERT} in the NLP field. Similar to the training method of BERT, the training of Spectrum-BERT is also divided into two stages: pre-training and fine-tuning. To retain the model's sensitivity to the characteristic peak position and local information of the spectral curve, we innovatively partition the curve into multiple blocks and obtain the embeddings of different blocks for the next calculation. 

In the pre-training stage, we elaborately design two pre-training tasks, 
\textbf{Next Curve Prediction (NCP)} and \textbf{Masked Curve Model (MCM)}, so that the model can effectively utilize unlabeled samples to capture the potential knowledge of spectral data, breaking the restrictions of the insufficient number of labeled samples. Spectrum-BERT, which adopts the "\normalem{\emph{pre-training \& fine-tuning" }} architecture, is different from the traditional tight coupling model of "feature extraction \& feature application": Spectrum-BERT separates latent feature extraction from different downstream tasks, reducing the dependency of downstream tasks on specific model architectures, and making the model more versatile and compatible.

To the best of our knowledge, \uline{this paper is the first attempt in the field to apply the "\normalem{\emph{pre-training \& fine-tuning}}" idea to spectral feature extraction and classification.}
The contributions of this paper can be summarized as the following 3 points:

\begin{itemize}

\item \textbf{The novel "\normalem{\emph{pre-training \& fine-tuning}}" paradigm.} We apply the idea of "\normalem{\emph{pre-training \& fine-tuning}}" to the field of spectral detection for the first time, and propose Spectrum-BERT. To retain the model's sensitivity to the characteristic peak position and local information of the spectral curve, we innovatively partition the spectral curve into multiple blocks and obtain the embeddings of different blocks for the next calculation.

\item \textbf{The elaborately designed pre-training tasks.} In the pre-training stage, we elaborately design two pre-training tasks, Next Curve Prediction (NCP) and Masked Curve Model (MCM), so that the model can effectively utilize unlabeled samples to capture the potential knowledge of spectral data, breaking the restrictions of the insufficient labeled samples, and improving the application ability and performance of the model in practical scenarios.

\item \textbf{The extensive experimental verification and analysis.} We conduct a large number of experiments on the real liquor spectral dataset. In the comparative experiments, the proposed Spectrum-BERT significantly outperforms the baselines in multiple metrics and this advantage is more significant on the imbalanced dataset. Moreover, in the parameter sensitivity experiment, we also analyze the model performance under different parameter settings, to provide a reference for subsequent related research.

\end{itemize}

\section{Related Work}

In this section, we will introduce and analyze the research work related to this paper, including spectral techniques for substance identification and the pre-training in deep learning.

\subsection{Spectral Techniques for Substance Identification}

With the advantages of easy operation, high sensitivity, low cost, and no pollution to the environment and samples, and long service life, spectral detection technology has been widely used in food detection. 
 Some researchers \cite{yan2021nondestructive,zhu2020deep,sheng2022analysis} combine spectroscopic techniques with classical machine learning for food identification and classification. Yan et al. \cite{yan2021nondestructive} leverage various principal component analysis (PCA) methods to extract features from the near-infrared hyperspectral image of peat barley malt, then input it into SVM to conduct classification, achieving high classification accuracy. To quickly determine the chemical components in milk, Sheng et al. \cite{sheng2022analysis} select a gradient-boosted regression tree (GBRT) for the NIR absorption multispectra of milk after completing the comparison of 8 machine learning models such as Logistic Regression, Random Forest, and Support Vector Regression (SVR). 

\begin{figure}
	\centering
	\includegraphics[width=3.5in,trim=20 20 10 15]{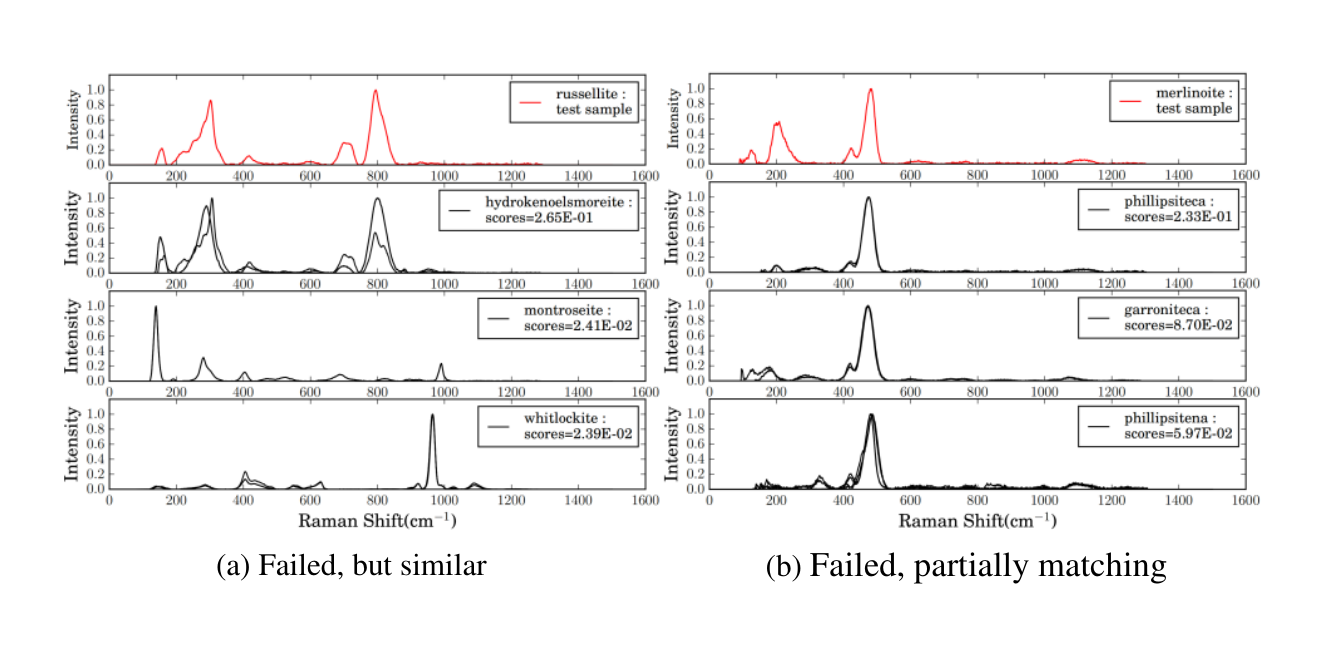}
	\caption{The shortcomings of CNN in extracting spectral curve features \cite{liu2017deep}. (a) Spectral curves with similar intensity characteristic peaks at different positions; and (b) spectral curves with similar characteristic peaks but mismatched details.}
	\label{FIG:2}
		\vspace{-1em}
\end{figure}

Some studies \cite{liu2021efficient,yan2021nondestructive,zhu2020deep} leverage Deep CNN as a solution for spectral recognition. 
For example, Zhu et al. \cite{zhu2020deep} employ two different methods of three-dimensional CNN (3-D CNN) and PCA combined with a two-dimensional CNN (2-D CNN), respectively, to conduct classification tasks on the near-infrared hyperspectral images from different kinds of candy and 2 different kinds of salmon fillet muscle tissue, and the classification performance is satisfactory. However, the previous study \cite{liu2017deep} has also shown that the Max-Pooling operation used in CNN may cause the model to lose local information in the process of downsampling. The spatial translation invariance of CNN also makes the captured spectral curve features lose the sensitivity to the position of characteristic peaks. Specifically, spectral curves with similar intensity characteristic peaks at different positions (as shown in Fig.\ref{FIG:2}(a)) or spectral curves with similar characteristic peaks but partially mismatched details (as shown in Fig.\ref{FIG:2}(b)) may be misjudged as similar curves by the model. This phenomenon is also confirmed in the experiments of \cite{yan2021nondestructive}: when classifying barley malt, Yan et al. \cite{yan2021nondestructive} show the classification performance of deep CNN and Auto-CNN respectively, and the results show that the performance of CNN is unsatisfactory. Therefore, although using CNN as the spectral curve recognition algorithm may achieve better accuracy in some cases \cite{liu2021efficient,zhu2020deep}, it can be seen from the studies of Liu et al. \cite{liu2017deep} and Yan et al. \cite{yan2021nondestructive} that, CNN has natural defects in extracting spectral curve features, and is not suitable as a feature extraction model for spectral curves.

The above studies use a combination of spectral technology and machine learning algorithms to conduct food identification or classification, but most of the used machine learning models are basic machine learning classification models (such as SVM, Random Forests, and Logistic Regression, etc.). On the one hand, the basic machine learning models used in the above methods have very limited feature capture and expression capabilities and high requirements for the quality of the input features. To meet the requirements of the model for input features, some researchers \cite{liu2021efficient,yan2021nondestructive,zhu2020deep} perform correlation analysis and extraction of high-dimensional spectral curves through algorithms such as PCA, wavelet transform (WT), and only curve segments with high correlation are chosen as input of the model. We believe that although this method achieves efficient classification of samples to some extent, it inevitably loses some detailed features, which will affect the recognition accuracy. On the other hand, most of the above-mentioned researchers construct spectral datasets by themselves. Since there is currently no large-scale public dataset on the liquor spectrum in the academic community, researchers need to construct and label liquor spectrum datasets. The cost is high, and the quality of the samples is difficult to guarantee. In practical application scenarios, there is not enough high-quality data to provide support for training efficient deep learning models.

Based on the above two considerations, it is a reasonable method to capture the rich features in the spectral curve by introducing a deep learning model with stronger expressive ability. However, deep learning models with strong expressive ability often require a large amount of labeled data for training. Therefore, how to effectively train large-scale deep learning models with limited spectral curve samples is a problem that needs to be addressed.

\subsection{Pre-training in Deep Learning.}

With the development of deep learning technology, machine learning models are developing towards larger model sizes and more complex computations \cite{howard2017mobilenets}. Therefore, more and higher quality samples are needed to support the training of more complex deep learning models. However, due to the limitation of sample acquisition difficulty and sample labeling cost, researchers dedicate themselves to solving few-shot learning problems. As the most effective scheme for few-shot learning, the "\normalem{\emph{pre-training \& fine-tuning}}" paradigm has achieved great success in NLP and CV \cite{vaswani2017attention,Bao0PW22,PetersNLSJSS19,carion2020end,BERT,DosovitskiyB0WZ21,radford2018improving,liu2019roBERTa,sun2019ernie}.

In the field of NLP, researchers \cite{2021Knowledge,JoshiCLWZL20,liu2019roBERTa,BERT,radford2018improving,sun2019ernie} have made an in-depth exploration of the \normalem{\emph{pre-training \& fine-tuning}} paradigm. 
BERT \cite{BERT} proposed by Devlin et al. utils the deep bidirectional Transformers encoder structure to solve the problem of understanding the meaning of words according to the context in the NLP field, from the perspective of model structure, and further improves the \normalem{\emph{pre-training \& fine-tuning}}. It has achieved ideal results on typical NLP problems, which provides a better idea for follow-up research. Many researchers have carried out rich exploratory studies based on the BERT architecture \cite{BERT}: Sun et al. \cite{sun2019ernie} extend the mask operation of a single word in BERT to words with complete word meanings, and propose an entity-based pre-training task, which improves the performance significantly; Joshi et al. \cite{JoshiCLWZL20} promote the model to learn longer-distance information by masking a whole piece of text, and the proposed SpanBERT achieves great performance in some segment extraction question-answering tasks.

On the other hand, many researchers \cite{carion2020end,chen2020generative,DosovitskiyB0WZ21,Bao0PW22,he2022masked} have also conducted in-depth research on the application of pre-trained models in the CV field. Dosovitskiy et al. \cite{DosovitskiyB0WZ21} propose the ViT model by constructing the image as a patch sequence as the encoder input, which pioneers the introduction of the Transformer-based model structure into the field of CV. But unfortunately, the ViT model is a supervised training model. Inspired by BERT \cite{BERT}, Chen et al. \cite{chen2020generative} propose iGPT, which creatively resizes the input image and then performs the mask operation, realizing an unsupervised training similar to BERT, which greatly reduces the training difficulty of the model. Bao et al. \cite{Bao0PW22} optimize iGPT \cite{chen2020generative}, leveraging the Visual Tokens output by the Tokenizer as a more accurate learning target for the Transformer, reducing the information loss caused by the resize operation in iGPT. 

The widespread application and outstanding performance of pre-training in the field of NLP and CV prove that the acquisition of general domain knowledge through pre-training can obtain efficient deep learning models.

\section{Spectrum-BERT: Pre-training of Deep Bidirectional Transformers for Spectral Classification of Chinese Liquors}

In this section, we will detail our proposed Spectrum-BERT model architecture, input and output representations, and the "\normalem{\emph{pre-training \& fine-tuning}}" procedure. As mentioned earlier, our proposed Spectrum-BERT is inspired by the BERT model proposed by Devlin et al. \cite{BERT}. BERT's deep bidirectional Transformers model and the idea of "\normalem{\emph{pre-training \& fine-tuning}}" have achieved outstanding achievements in the fields of NLP and CV. Still, no researcher has applied it to the area of spectral feature extraction and classification. BERT's characteristic of leveraging unlabeled data for training in the pre-training phase is the main reason why we choose to transfer the "\normalem{\emph{pre-training \& fine-tuning}}" to the domain of spectral classification. Moreover, we believe that the \normalem{\emph{Position Embedding}} introduced by BERT in the training process can enable the model to capture the characteristic peak and local information of the spectral curve during learning, so as to deal with the problem that the CNN loses the characteristic peak position information when extracting the spectral curve features \cite{liu2017deep}.

Referring to the "\normalem{\emph{pre-training \& fine-tuning}}" of BERT \cite{BERT}, Spectrum-BERT also consists of two training steps: pre-training and fine-tuning. First, Spectrum-BERT is pre-trained (section \ref{sec3.3}) using the unlabeled spectral data of liquor processed by the data preprocessing method (section \ref{sec4.2}). Then, the pretrained Spectrum-BERT is differentially fine-tuned for a custom downstream task (section \ref{sec3.4}). 


\subsection{Model Architecture}\label{sec3.1}

\begin{figure}[t]
	\centering
	\includegraphics[width=3.5in]{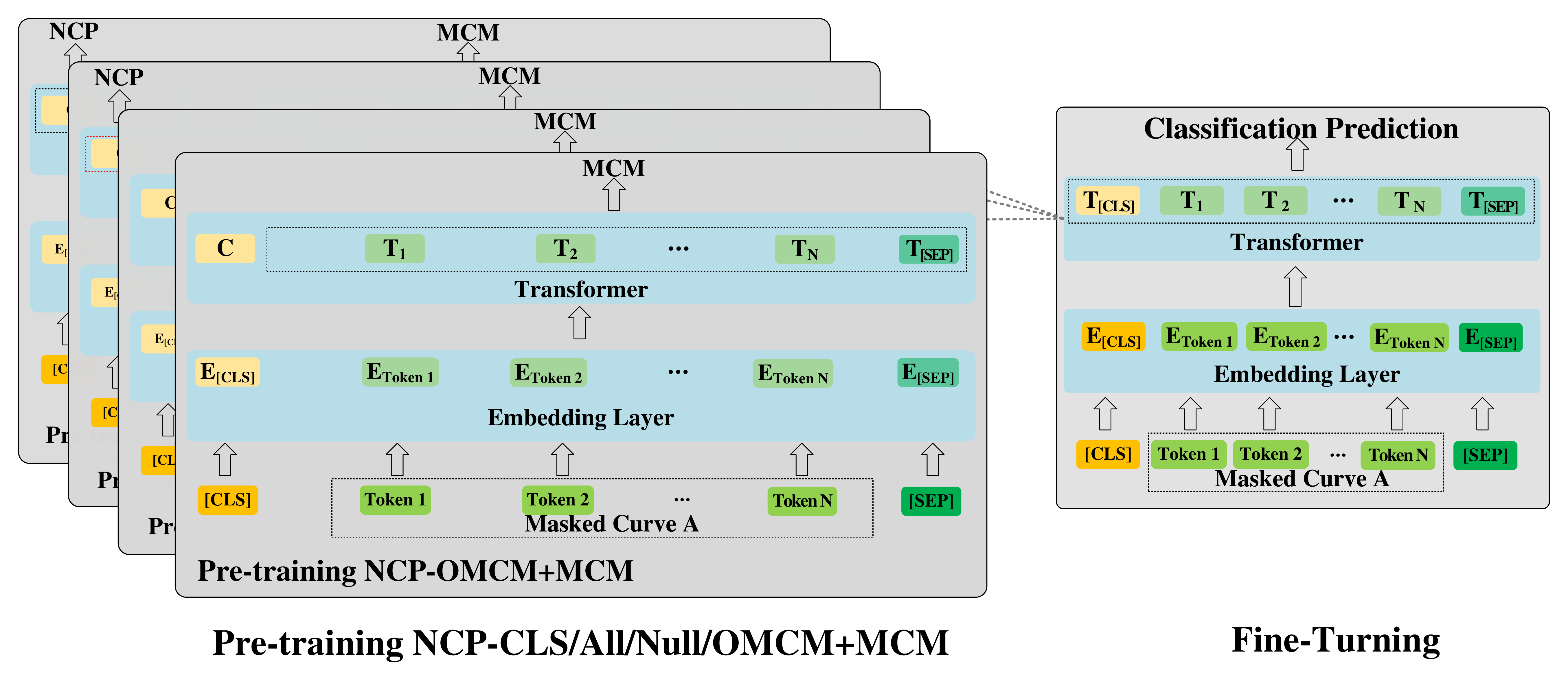}
	\caption{The overall process of pre-training and fine-tuning of Spectrum-BERT. The two stages are identical except for the optimization tasks. The parameters of Spectrum-BERT in the fine-tuning stage are initialized using the parameters obtained in the pre-training stage, and all parameters will be fine-tuned. The token [CLS] is the start token of the input sequence, and the token [SEP] is the split token and end token in the input sequence. The four different pre-training tasks (Pre-training NCP-CLS/All/Null/OMCM+MCM) will be introduced in detail in section \ref{sec3.3}.}
	\label{FIG:3}
	\vspace{-1em}
\end{figure}

As shown in Fig.\ref{FIG:3}, the overall architecture of Spectrum-BERT is composed of deep bidirectionally connected Transformers, which can be divided into the input layer (Embedding Layer), the model layer (Transformer), and the downstream task layer. The calculation method and model architecture of the input layer will be introduced in section \ref{sec3.2}. The Spectrum-BERT size constructed by the model layer is mainly defined by the following three parameters: The number of layers $L$ (i.e. the number of stacked Transformer blocks in the model layer), the hidden layer size $H$ (i.e. the size of each token block output by Spectrum-BERT), and the number of self-attention heads $A$. 
The main role of the downstream task layer is to use the feature representation output by the model layer to perform specific tasks (sections \ref{sec3.3} and \ref{sec3.4}). Specifically, Spectrum-BERT takes our well-designed Next Curve Prediction (NCP) task and Masked Curve Model (MCM) task as training targets in the pre-training stage and uses a multi-classifier built with fully connected layers for the multi-classification task in the fine-tuning stage. Note that although the training tasks performed in the two stages are different, the feature representations used are both spectral feature representations from the model layer and do not involve differences in the overall architecture of Spectrum-BERT.

\subsection{Spectral Curve Partition and Input Layer}\label{sec3.2}

Due to the continuity of the spectral curves, we cannot process the spectral curves using methods similar to the pre-trained word embedding models[26] in NLP to obtain \normalem{\emph{Token Embeddings}}. Therefore, to retain the model's sensitivity to the characteristic peak position and local information of the spectral curve, we innovatively partition the spectral curve into multiple blocks. The hyperparameter of $token\_size$ determines the length of the block. Then 1-D convolution is used to map the block (we call it token) into \normalem{\emph{Token Embeddings}} with length $H$ (the preset hidden size), which is used as feature input for the subsequent calculation. Specifically, we define the feature mapping process using:

\begin{equation}
E_{Token_{i}}=trans\left( Token_{i}\right)  =Cov1d\left( Token_{i}\right)  
\end{equation}
where $Token_i$ is the $i$-th in the token sequence, $E_{Token_i}$ is the \normalem{\emph{Token Embeddings}} corresponding to the $i$-th token, $i\in [1,A_n+B_n]$, $A_n$($B_n$) is the number of curve $A$($B$) tokens. Specifically, 
the original spectral curve input to the input layer will first be partitioned into $token\_size$-dimensional tokens, and a 1-D convolutional layer converts $token\_size$-dimensional tokens into $H$-dimensional \normalem{\emph{Token Embeddings}}.

\begin{figure}[t]
	\centering
	\includegraphics[width=3.5in]{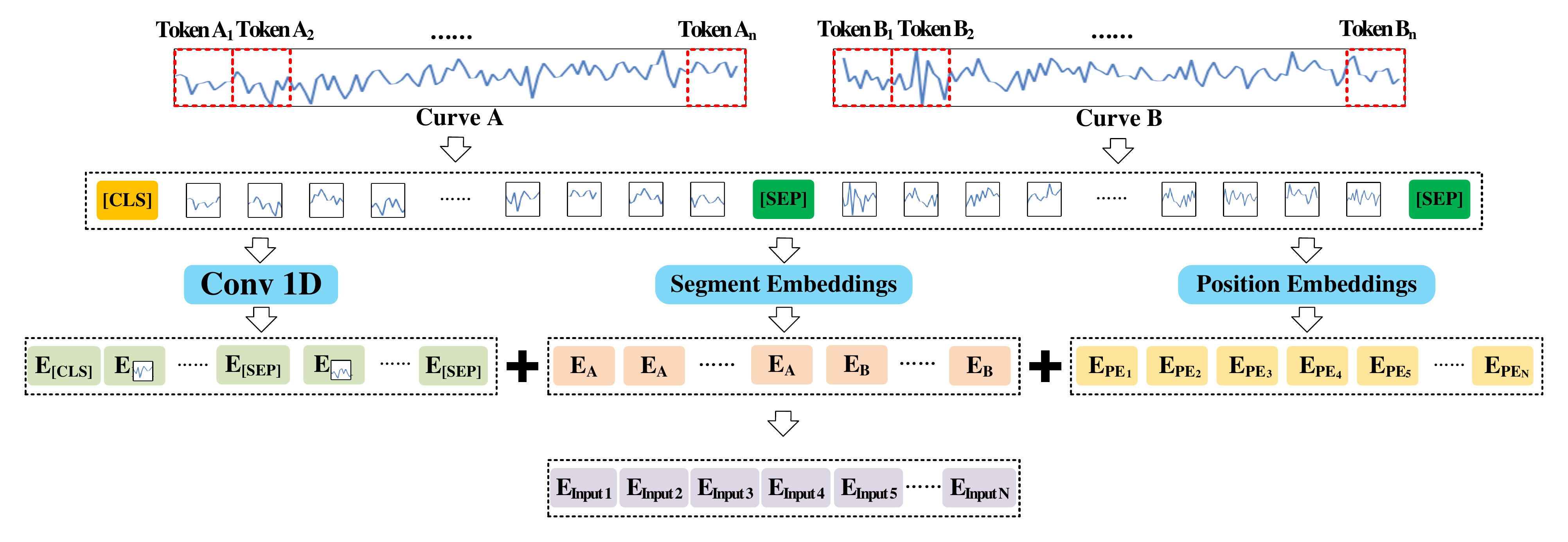}
	\caption{The diagram of the Spectrum-BERT input layer. The input feature is obtained by summing three parts: \emph{Token Embedding}, \emph{Segment Embedding}, and \emph{Position Embedding}.}
	\label{FIG:4}
	\vspace{-1em}
\end{figure}

To accomplish specific pre-training or fine-tuning tasks, sometimes we need to input a pair of two spectral curves into Spectrum-BERT as a sequence of tokens. We adopt two measures of special marker segmentation and \normalem{\emph{Segment Embedding}} to distinguish the two curves. First, we insert the [SEP] token as a delimiter between the two curves and at the end of the entire token sequence (for single-curve input, the [SEP] token is only at the end of the token sequence). Second, we add learnable \normalem{\emph{Segment Embeddings}} to each token feature representation to indicate that the current token block comes from either curve $A$ or curve $B$. As shown in Fig.\ref{FIG:4}, we use $E_A$ to indicate that the current token block comes from curve $A$, and $E_B$ to indicate that the current token block comes from curve $B$. For the design of \normalem{\emph{Position Embeddings}}, we follow the idea of Vaswani et al. \cite{vaswani2017attention} and leverage the linear transformation of $sin$ and $cos$ functions to add position information to each spectral curve block:
\begin{equation}
E^{k}_{PE_{i}}=\begin{cases}sin\left( \frac{pos}{1000^{\frac{2d}{H} }} \right)  ,&d=2k\\ cos\left( \frac{pos}{1000^{\frac{2d}{H} }} \right)  ,&d=2k+1\end{cases} 
\end{equation}
where $E_{PE_i}^k$ is the $k$-th dimension in \normalem{\emph{Position Embeddings}} corresponding to the $i$-th token, $pos\in [0,max\_seq\_length)$ is the position number of the token in the token sequence, $k\in [0,H/2)$ is the dimension number in \normalem{\emph{Position Embeddings}}, $H$ is the hidden layer size, $max\_seq\_length$ is the maximum length of the token sequence. In this way, Spectrum-BERT can learn the location information of characteristic peaks in the spectral curve. Finally, we sum the three parts (\normalem{\emph{Token Embeddings}}, \normalem{\emph{Segment Embeddings}}, and \normalem{\emph{Position Embeddings}}) corresponding to the same token block to get the final input representation sequence:
\begin{small}
	\begin{equation}
E_{Input_{i}}=\begin{cases}E_{Token_{i}}+E_{A}+E_{PE_{i}},&if\  Token_{i}\  from\   A\\ E_{Token_{i}}+E_{B}+E_{PE_{i}},&if\ Token_{i}\  from\    B\end{cases}  
\end{equation}
\end{small}
where $E_{Input_{i}}$ is the final input representation corresponding to the $i$-th token in the input token sequence, $E_{PE_{i}}$ is the \normalem{\emph{Position Embeddings}} corresponding to the $i$-th token.

We have described the model architecture and calculation method of the input layer in Spectrum-BERT in detail. It should be noted that the beginning of each token sequence processed by the input layer is a [CLS] mark. Meanwhile, the output $C\in \mathbb{R}^{H}$ corresponding to this token represents the aggregated representation of the entire sequence of input tokens. Similarly, we define the feature representation corresponding to the $i$-th curve block behind the [CLS] marker as $T_i \in \mathbb{R}^{H}$.

\subsection{Pre-training Spectrum-BERT}\label{sec3.3}

In the pre-training stage, we design the following two pre-training tasks:

\textbf{Task \#1: Masked Curve Model (MCM)}. 
We expect that the model can infer or predict the unknown missing content based on the known context like a human, but simply connecting two unidirectional models cannot reach the expected goal. Therefore, we design the Masked Curve Model (MCM) pre-training task. Specifically, each token in the input sequence of tokens processed by the input layer will be masked with a 15\% probability. The feature representation corresponding to the masked token output by the model layer needs to be as similar to the token embedding before being masked as possible. We define the loss function $L_{MCM}$ for the MCM task as:

\begin{equation}
L_{MCM}=\frac{1}{|S|} \sum^{}_{i\in S} \left( T_{i}-Token_{i}\right)^{2}   
\end{equation}
where $S$ is the serial number set of the masked token. In the MCM task, we expect the output of the Spectrum-BERT to approximate the masked token as closely as possible, rather than reconstructing the entire input token sequence. This measure avoids the problem of information leakage in traditional deep bidirectional learning and enables the model to predict missing tokens from two directions (before and after the current position) of the curve, thereby capturing the compositional characteristics or composition patterns of the entire input token sequence.

Please note that we do not need to replace all masked tokens with the [MASK]. When a token in the input sequence needs to be masked, there are three different strategies: (1) 80\% probability to be replaced by [MASK] (total probability is 15\%×80\%=12\%); (2) 10\% probability to be replaced by a random token (total probability is 15\%×10\%=1.5\%); (3) 10\% probability to keep the original token unchanged (total probability is 15\%×10\%=1.5\%). We design masking strategies for MCM tasks from the perspective of narrowing the training gap between pre-training and fine-tuning. If each selected token is masked with a fixed probability in the pre-training stage, the model will completely lose the ability to perform prediction and reasoning based on the semantics of the masked token, which is detrimental to the feature extraction and semantic expression capabilities of the entire model. We expect that Spectrum-BERT can take the semantics of the masked token into account while trying to learn the semantic information around and distant from the masked token so that Spectrum-BERT has the ability to model the complete semantic information of the curve.
\begin{figure}[t]
	\centering
	\includegraphics[width=3.5in]{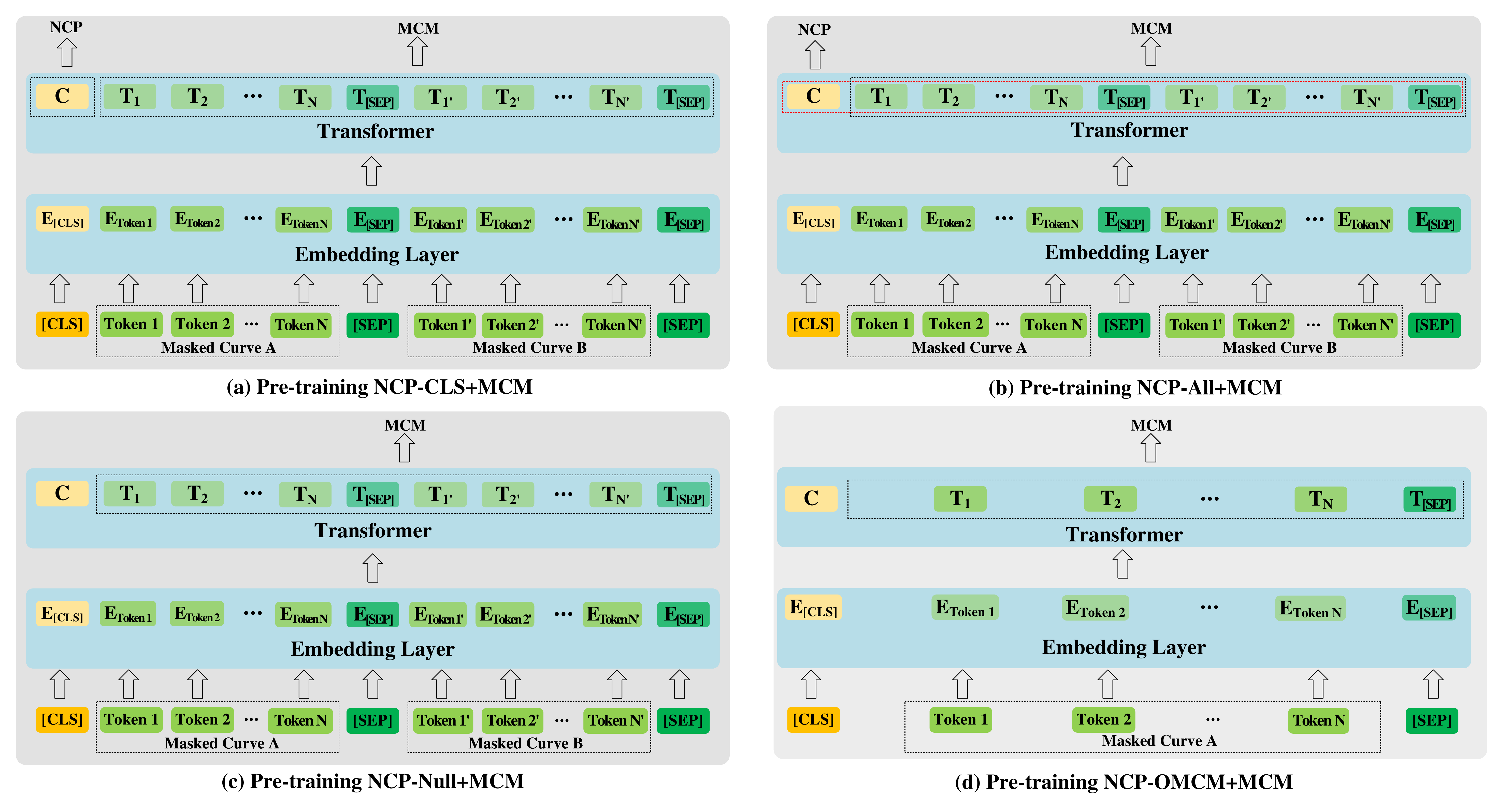}
	\caption{The schematic diagram of the detailed structure of the four pre-training task models.}
	\label{FIG:5}
	\vspace{-1.2em}
\end{figure}

\textbf{Task \#2: Next Curve Prediction (NCP)}. In the field of NLP, we expect a good language model to capture the relationship between sentences while performing language modeling. Similarly, we expect that Spectrum-BERT can capture the correlation between different curves while modeling the characteristics of spectral curves. Therefore, we propose the Next Curve Prediction (NCP) task and its variants for spectral curves, named NCP-CLS, NCP-All, NCP-Null, and NCP-OMCM.

As shown in Fig.\ref{FIG:5}(a), in the NCP-CLS pre-training task, we take the curve pair composed of two spectral curves as the input of Spectrum-BERT, and the aggregate representation $C\in \mathbb{R}^{H}$ corresponding to [CLS] is used as the input of the NCP task in the downstream task layer. In the downstream task layer, we employ the SoftMax function for binary classification tasks:

\begin{equation}
\hat{y} =softmax\left( C\right)     
\end{equation}
where $\hat{y} $ represents the result of the binary classification task. In the process of the experiment, we adopt the method of random combination to obtain curve pairs. Specifically, there is a 50\% probability that spectral curve $A$ and $B$ belong to the same liquor (the label is 1), and there is a 50\% probability that spectral curve $A$ and $B$ belong to different liquors (the label is 0). We define the loss function $L_{NCP-CLS}$ for the NCP-CLS task as:

\begin{equation}\label{eq6}
{}L_{NCP-CLS}=-\frac{1}{B} \sum^{B}_{i=1} log(\hat{y}^{i}_{y_{i}} ),\  y_{i}\in \left\{ 0,1\right\}  
\end{equation}
where $B$ is the batch size, $y_i$ is the label of the $i$-th sample, ${}\hat{y}^{i}_{j}$ is the probability that the model predicts the $i$-th sample as belonging to class $j$. The pre-training loss function $L_{NCP-CLS+MCM}$ for Spectrum-BERT using the NCP-CLS pre-training task is expressed as:

\begin{equation}
L_{NCP-CLS+MCM}=L_{NCP-CLS}+L_{MCM} 
\end{equation}

Considering that only leveraging the aggregate representation of the input sequence as the feature for NCP tasks may be insufficient, we propose the NCP-All pre-training task. As shown in Fig.\ref{FIG:5}(b), in the NCP-All pre-training task, we input the aggregate representation $C \in \mathbb{R}^{H}$ and the entire feature representation sequence $T_i \in \mathbb{R}^{H}$ to the downstream task layer for the NCP pre-training task after concatenation operation. In the NCP-All pre-training task, the classification result of the classifier is expressed as:
\begin{equation}
\hat{y} =softmax\left( C\oplus T_{i}\right)  ,i=1,2,...,N
\end{equation}
where $\oplus$ is the concatenation operation on vectors $C$ and $T_i$, $N$ is the length of the input token sequence. The definition of the loss function $L_{NCP-All}$ for the NCP-All pre-training task is consistent with Equation (\ref{eq6}). The pre-training loss function $L_{NCP-All+MCM}$ for Spectrum-BERT using the NCP-All pre-training task is expressed as:

\begin{equation}
L_{NCP-All+MCM}=L_{NCP-All}+L_{MCM}
\end{equation}

The original intention of the Next Sentence Prediction (NSP) pre-training task proposed by Devlin et al. \cite{BERT} is to solve problems such as Question and Answer (QA) and Natural Language Inference (NLI) in the field of NLP, while this problem does not exist in the field of spectral classification. At the same time, we believe that there may be an optimization conflict between the NCP and MCM tasks in the NCP pre-training. Based on the above two viewpoints, we propose the NCP-Null pre-training task. As shown in Fig.\ref{FIG:5}(c), in the NCP-Null pre-training task, the composition of the model input is the same as in the NCP-All and NCP-CLS pre-training tasks, but the backpropagation for NCP pre-training tasks is not performed. Specifically, the pre-training loss function $L_{NCP-Null+MCM}$ of Spectrum-BERT using the NCP-Null pre-training task can be expressed using the following equation:

\begin{equation}
L_{NCP-Null+MCM}=L_{MCM}
\end{equation}

The input of the NCP-Null pre-training task is consistent with the NCP-All and NCP-CLS pre-training tasks, but whether the two curves that make up the curve pair belong to the same liquor is not distinguished by the NCP task. Therefore, we believe that the NCP-Null pre-training task makes the spectral curve characteristics of liquor captured by Spectrum-BERT biased. To resolve the above problem, we propose the NCP-OMCM pre-training task. As shown in Fig.\ref{FIG:5}(d), in the NCP-OMCM pre-training task, the input of Spectrum-BERT is a single curve rather than a pair of two curves, and other model details are consistent with the NCP-Null. Specifically, we define the loss function $L_{NCP-OMCM+MCM}$ for the NCP-OMCM pre-training task using the following equation:

	\begin{equation}
	\begin{aligned}
L_{NCP-OMCM+MCM}&=L^{\prime }_{MCM}\\&=\frac{1}{|S^{\prime }|} \sum^{}_{i\in S^{\prime }} \left( T_{i}-Token_{i}\right)^{2}  
		\end{aligned}
\end{equation}
where $S^{\prime }$ is the serial number set of the masked token from a single curve.

\subsection{Fine-tuning Spectrum-BERT}\label{sec3.4}

The primary purpose of the fine-tuning stage is to adjust the model parameters obtained from the pre-training stage to better match downstream tasks in different scenarios. Due to the universality of the model structure of Spectrum-BERT, without changing the model architecture, it is only necessary to replace the downstream task layer to complete specific tasks in various scenarios (for example, classification, matching, clustering tasks, etc.).

In this paper, we take the classification task of the liquor spectral curve as the target task for fine-tuning training. To narrow the training gap between the fine-tuning and pre-training stages, we keep the processing of model inputs in the fine-tuning stage consistent with the pre-training stage. It should be noted that the parameter $token\_size$ in the fine-tuning stage needs to be consistent with that in the pre-training stage, but the model input in the fine-tuning stage is a single curve instead of curve pairs. We explore two different pre-training tasks in the fine-tuning stage, namely spectral curve classification based on [CLS] tokens and spectral curve classification based on all tokens. In the spectral curve classification task based on [CLS] tokens, we take the aggregate representation ($C\in \mathbb{R}^{H}$) corresponding to the [CLS] token as the input of the linear classifier (this is also the scheme of BERT). In the spectral curve classification task based on all tokens, we take the entire feature representation sequence ($T_i \in \mathbb{R}^{H}$) as the input of the linear classifier. After experimental verification, we found that the performance of using the [CLS] token for spectral curve classification is unsatisfactory. Therefore, in the experiment section(see section \ref{sec4}), we prefer to the spectral curve classification task based on all tokens.


\section{Experiment}\label{sec4}

In this section, we first introduce a brief description of the experiment, including the experiment instrument, datasets and preprocessing methods, baseline model definition, and basic parameter settings. Second, we demonstrate the contribution of this paper by comparing Spectrum-BERT and its variants with baselines, and further demonstrate the effectiveness of Spectrum-BERT through exploration experiments. Finally, we conduct sensitivity experiments on some important hyperparameters such as $token\_size$, self-attention heads ($A$), and hidden layer size ($H$).

\subsection{Experiment Instrument}\label{sec4.1}

\begin{figure}
	\centering
	\includegraphics[width=3in,trim=0 20 0 0]{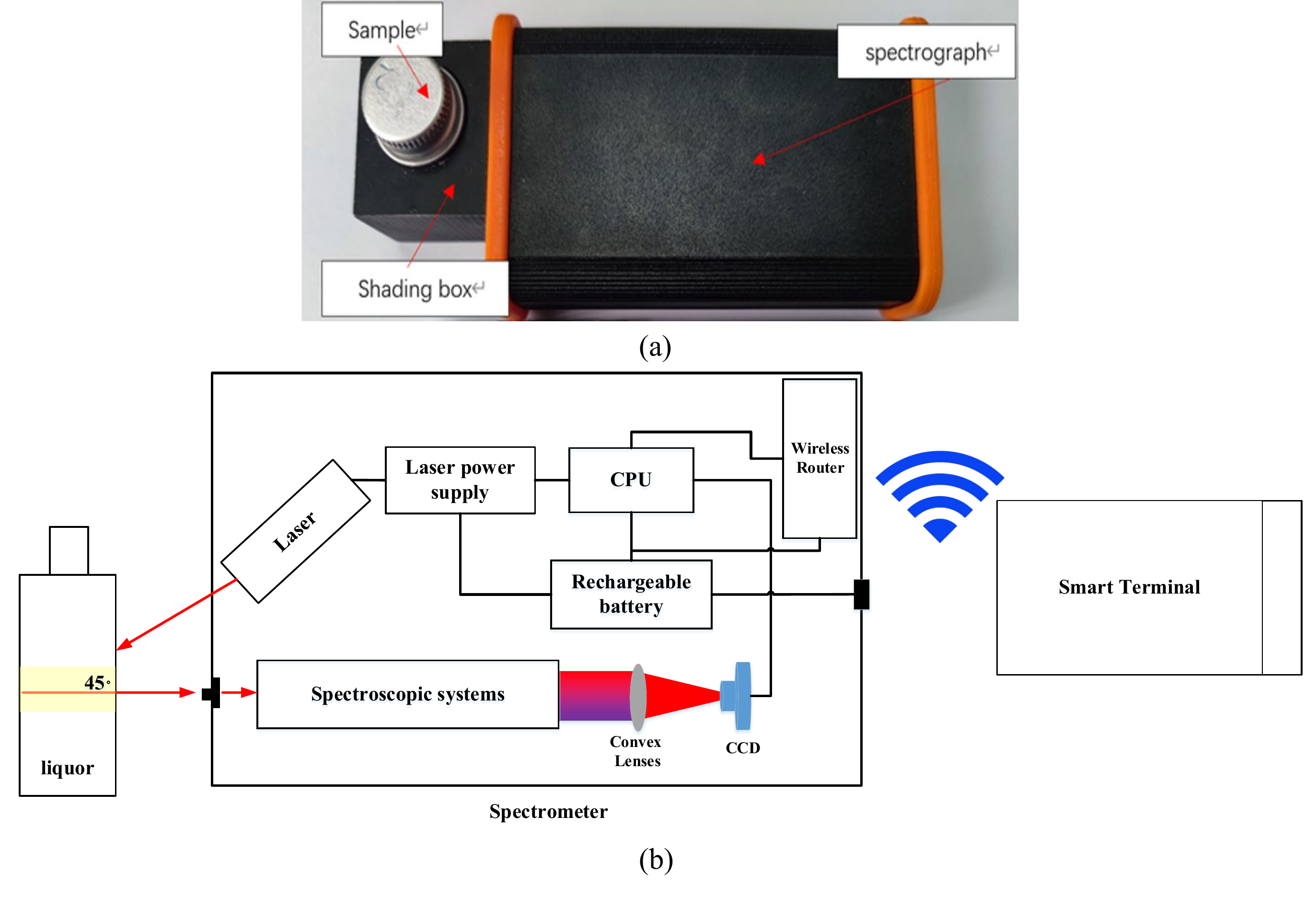}
	\caption{(a) The physical diagram and (b) the internal structure of experiment systems.}
	\label{FIG:6}
		\vspace{-1.3em}
\end{figure}

The experimental system is shown in Fig.\ref{FIG:6}(a). The size of the micro-spectrometer is 103mm×58mm×25mm, which is our self-developed. The wavelength of the laser diode is 405 nm and we can switch output mode between continuous or pulse output mode as required. The CCD pixel is 1280×720, and the CPU chip is Atheros AR9331 SOC. The internal structure of the experimental system is shown in Fig.\ref{FIG:6}(b), which includes a laser spectrometer and a smart terminal. The laser spectrometer consists of a CPU, a spectroscopic system, convex lenses, and a battery. The CPU is respectively connected to the laser emission module, the CCD, and the wireless routing module. The laser emitted from the laser emission module excites the tested liquor through the wall of the bottle, and the optical signal generated by the spectroscopic system is then concentrated on the surface of the CCD by the convex lens. The CCD can convert the optical signal into an electrical signal, the CPU is responsible for receiving the control instructions of the smart terminal and controlling the laser emission module to emit laser with preset laser intensity, and then converting the electrical signal transmitted by the CCD into a digital image. The wireless router module undertakes the task of wireless communication between the smart terminal and the micro-spectrometer. When conducting experiments, the liquor samples are put in a clear glass bottle and placed into the shading box for testing. The smart terminal receives spectral data through the wireless router.

\begin{table*}[t]
  \setlength{\abovecaptionskip}{-0.2cm} 
  \caption{The results of the comparative experiment on the spectral classification task (\%). SVM and RF do not belong to deep neural networks, so the number of model parameters and calculation amount cannot be counted. Bold for “the best”, and underline for “the second best”.}
  \label{tab1}
  \setlength\tabcolsep{2.2pt}
  \renewcommand{\arraystretch}{1.3}
  \scriptsize
  \center
  \begin{tabular}{@{}cccccc|cccc@{}}
  \toprule
  Metrics		& Train ratio
   & RF		& BP  	 & SVM
  & 1-D-Deep-CNN				& $\text{Spectrum-BERT}_{\text{CLS}}$   & $\text{Spectrum-BERT}_{\text{Null}}$   	& $\text{Spectrum-BERT}_{\text{All}}$
  & $\text{Spectrum-BERT}_{\text{OMCM}}$\\ \midrule 

   \#Params(M)	& -  
  		  & -		& 0.26   & -
  & 68.13 		& 3.22 		  & 3.22     	& 3.22 
  & 3.22  \\ \midrule 
  
  \#FLOPs(M)	& -
    		& -		& 1.05  & -	
  & 47774.41 	& 4469.92 	  & 4469.92   & 4469.92 
  & 4469.92  \\ \midrule 
  
	\multirow{4}{*}{Precision} 	 		& 64\%     
  & $56.28\pm{27.03}$			& $85.41\pm{50.98}$  & $94.61\pm{0.99}$		 	
  & \bm{$99.89\pm{0.05}$	}			& $97.85\pm{7.71}$ 		  	& $98.67\pm{0.08}$  & $98.43\pm{0.25}$ 
  &  \underline{$98.76\pm{0.25}$}       \\ 

  &                           42\%  
  & $64.22\pm{12.51}$		& $84.03\pm{34.55}$  & $84.85\pm{34.78}$		  
  & \bm{$99.49\pm{0.10}$	} 			& $97.10\pm{0.59}$ 		  &$97.22\pm{1.08}$  & \underline{$97.48\pm{0.19}$}
  &  $97.40\pm{0.99}$      \\
  
  &                           25\%
  & $61.83\pm{14.47}$		& $87.44\pm{21.04}$  & $85.39\pm{34.78}$		  
  & $97.01\pm{1.35}$				& $96.63\pm{0.62}$ 	 &$96.96\pm{0.76}$   & \underline{$97.12\pm{0.70}$}
  &  \bm{$97.27\pm{0.26}$}       \\
  
  &                            16\%        
 & $55.56\pm{6.03}$		& $88.44\pm{44.56}$   & $85.45\pm{0.45}$		  
  & $95.99\pm{1.52}$				& $95.98\pm{1.05}$ 	&$95.61\pm{0.93}$  & \underline{$95.98\pm{0.86}$} 
  &  \bm{$96.11\pm{0.42}$}      \\ 
  
  \midrule

  \multirow{4}{*}{Recall} 
  & 64\%
  & $60.88\pm{20.74}$		& $89.27\pm{36.94}$   & $92.65\pm{1.52}$		 
  & \bm{$99.88\pm{0.06}$	}			& $97.60\pm{8.19}$ 	&$98.49\pm{0.08}$  	  & $98.13\pm{0.58}$ 
  &  \underline{$98.62\pm{0.28}$}      \\ 
          
  & 42\%       
  & $67.89\pm{7.45}$		& $88.46\pm{26.63}$  & $88.46\pm{9.52}$		  
  & \bm{$99.45\pm{0.14}$}				& $96.47\pm{0.52}$ &$96.74\pm{1.06}$  		  & $96.83\pm{0.37}$ 
  &  \underline{$97.01\pm{1.10}$}      \\
    
                            & 25\%     
  & $62.66\pm{9.37}$		& $90.78\pm{9.52}$  & $90.42\pm{0.25}$		  
  & $95.62\pm{3.74}$				& $96.12\pm{0.81}$  &$96.27\pm{0.85}$  		 & \underline{$96.74\pm{0.75}$}
  &  \bm{$96.78\pm{0.33}$}      \\
     
                          & 16\%    
  & $54.41\pm{9.68}$		& $31.32\pm{24.28}$  & $90.50\pm{0.50}$		  
  & $93.91\pm{6.09}$				& $94.64\pm{1.33}$ &$94.61\pm{1.03}$   	 & \underline{$94.70\pm{1.34}$}
  &  \bm{$95.24\pm{0.85}$}      \\ \midrule 
  
\multirow{4}{*}{weighted-F1} 
  
  & 64\%
  & $56.36\pm{22.98}$		& $86.57\pm{47.26}$  & $91.32\pm{2.70}$		  		
  & \bm{$99.88\pm{0.06}$	}			& $97.58\pm{8.28}$  &$98.48\pm{0.08}$  		  & $98.07\pm{0.74}$
  &  \underline{$98.61\pm{0.27}$}      \\ 
          
  & 42\%       
  & $64.04\pm{9.32}$		& $85.31\pm{34.89}$ & $84.58\pm{19.06}$		  		
  & \bm{$99.45\pm{0.14}$}				& $96.47\pm{0.56}$ &$96.78\pm{1.05}$   		  & $96.83\pm{0.38}$
  &  \underline{$97.03\pm{1.10}$}      \\
    
  & 25\%     
  & $59.68\pm{11.04}$		& $87.98\pm{13.02}$ & $87.34\pm{0.25}$		  		
  & $95.23\pm{4.82}$				& $96.10\pm{0.81}$ &$96.63\pm{0.92}$    & \underline{$96.74\pm{0.73}$}
  &  \bm{$96.77\pm{0.35}$}      \\
     
  & 16\%    
  & $51.11\pm{11.95}$		& $88.94\pm{34.35}$  & $87.43\pm{0.52}$		  	
  & $93.30\pm{9.69}$				& $94.75\pm{1.33}$ 	 &$94.64\pm{1.04}$  	& \underline{$94.78\pm{1.31}$}
  &  \bm{$95.33\pm{0.79}$}      \\ \midrule 
    
  \vspace{-0.8cm}
  \end{tabular}
  \end{table*}
\subsection{Dataset}\label{sec4.2}

The tested liquors are selected from 12 brands in the Chinese market, including 3 Maotai-Flavor liquors, 3 Luzhou-Flavor liquors, and 6 Fen-Flavor liquors. The names and numbers are Beijing Erguotou (\#1), Weihai Wei Shao Guo (\#2), Clove Love (\#3), Raw Pulp (\#4), Beidacang (\#5), Bitter Mustard (\#6); Jing Zhi Bai Gan (\#7), Hengshui Lao Bai Gan (\#8), Lao Jiu Hu (\#9), Niu Lan Shan (\#10), Du Er Jiu (\#11) and Xiao Lang Jiu (\#12). 100 sample data were taken at different times for each liquor. To ensure the accuracy and consistency of the measurement results, we first calibrate the laser intensity of the instrument, and the sample used for calibration is pure water. To achieve the background deduction, the data under the condition of laser on and off is subtracted to obtain the spectral curve. In this way, the average value of five times is taken as the final data to eliminate errors. Finally, we normalize the spectral data using the Min-Max normalization method to eliminate the influence of scale on subsequent calculations.

\subsection{Experiment Setup}\label{sec4.3}

\subsubsection{Baselines and Model Definition} We select some representative basic machine learning models and deep learning models as the baselines for comparison. At the same time, to verify the effect of different modules, we also construct different variants of our model.

\begin{itemize}

\item SVM \cite{yan2021nondestructive}: 

Following \cite{yan2021nondestructive}, the SVM is used for spectral curve classification. After testing with our liquor spectrum dataset, we set the parameter settings of SVM in the experiment as follows: the kernel function is RBF, $C=1$, and $\gamma =1/(n\_features*X.var)$, where $n\_features$ is the feature dimension, $X.var$ is the data variance.

\item RF \cite{sheng2022analysis}:

Following Sheng et al. \cite{sheng2022analysis}, we use the grid search algorithm to tune the Random Forest (RF) model. Specifically, we search for $n\_estimators$ (the number of trees) in the range of [20, 140] with the step size 20, and $max\_depth$ (the depth of a single decision tree) in the range of [1, 20] with the step size 5, respectively. And finally, $n\_estimators$ is 110 and $max\_depth$ is 6.

\item BP\cite{sheng2022analysis}

We refer to the backpropagation neural network (BPNN) architecture used in the comparative experiments by Sheng et al. \cite{sheng2022analysis}, and after testing with our liquor spectrum dataset, we determine to use a BP network with a hidden layer of 2 to classify liquor spectrum curves. The number of neurons in the input layer, output layer, and two hidden layers is 1000, 13, 250, and 50. After a lot of experiments, we determine that the batch size is 64, the learning rate is 0.001, and the maximum iteration is 1000. At the same time, we apply the early stopping mechanism (patience value is 20) in the experiment.

\item 1-D-Deep-CNN \cite{sang2022one}

We apply the one-dimensional convolution-based deep CNN model (1-D-Deep-CNN) proposed by Sang et al. \cite{sang2022one} to the task of liquor spectral curve classification. We follow the practice of Sang et al. \cite{sang2022one} to build a 1-D-Deep-CNN using five convolutional blocks and two fully connected blocks. Referring to the original hyperparameter settings of 1-D-Deep-CNN, we make appropriate adjustments according to the actual situation and finally determine: the learning rate is 0.001, the batch size is 64, and the maximum iteration is 200. At the same time, we apply early stopping in the experiment (patience value is 30). 

\item $\text{Spectrum-BERT}_{\text{CLS/All/Null/OMCM}}$:

According to the Spectrum-BERT pre-training task description given in section \ref{sec3.3}, we use NCP-CLS, NCP-All, NCP-Null, and NCP-OMCM as the pre-training task to obtain various variants of Spectrum-BERT for experiments, abbreviated as $\text{Spectrum-BERT}_{\text{CLS}}$, $\text{Spectrum-BERT}_{\text{All}}$, $\text{Spectrum-BERT}_{\text{Null}}$, and $\text{Spectrum-BERT}_{\text{OMCM}}$, respectively.

\end{itemize}

\subsubsection{Basic settings} The parameters of the Spectrum-BERT proposed in this paper are initialized by random initialization, and the Adam optimizer \cite{kingma2014adam} is used to optimize the model. The learning rate of Adam is set to 0.001, weight decay is 0.01, $\beta_{1}=0.9$, $\beta_{2}=0.999$. 
We adopt the early-stop strategy in the training process, and the patience is 20, i.e., we stop training if the validation performance does not increase for patience consecutive 20 epochs, the $max\_epoch=2000$. We leverage grid search strategy to determine the optimal values of hyperparameters of $L$, $A$, $H$, and $token\_size$ in \{4,8,12\}, \{4,8,16\}, \{256,768,1024\}, and \{50,100,200\}, and the optimal values are finally determined as $L=8$, $A=8$, $H=256$, $token\_size=100$. We partition the dataset into  $train:valid:test=(1-test\_rate)^2:test\_rate×(1-test\_rate):test\_rate$ with hyperparameter $test\_rate$ (0.2 by default). In the pre-training stage, batch size is 64. In the fine-tuning stage, the batch size is 128. Without the special declaration, the experimental parameters are set according to the above instructions. To ensure the stability of the experimental results, all the results in this paper are the average of 10 experimental results and the variances are recorded.

\subsubsection{Metrics} We use three metrics, Precision, Recall, and F1 score (weighted-F1) to evaluate model performance comprehensively. To fully consider the uneven distribution of samples of different categories in the test data, the above three metrics are calculated by the weighted average of the sample proportion with various labels.

\subsection{Comparative Experiment}\label{sec4.4}

We conducted comparative experiments on the model according to the experimental settings defined in section \ref{sec4.3} The final results are shown in Table \ref{tab1}. According to the experimental results of Table \ref{tab1}, we can draw the following conclusions:

\begin{enumerate}[(1)]

\item Our proposed method achieves the best or most competitive performance on all metrics with fewer parameters and computational budgets. With sufficient training samples (such as 64\% or 42\%), the 1-D-Deep-CNN proposed by Sang et al. \cite{sang2022one} can achieve excellent performance, but it needs 10.7 times parameters and 5.4 times computational budgets compared to our models. $\text{Spectrum-BERT}_{\text{OMCM}}$ achieves comparable performance to 1-D-Deep-CNN with a relatively small amount of model parameters and computational budgets. This proves that our proposed Spectrum-BERT model architecture achieves efficient latent feature capture and knowledge representation with fewer model parameters. Compared with other machine learning models,  our proposed Spectrum-BERT model has more excellent and stable performance, which indicates that the Spectrum-BERT model is more suitable for capturing the position-sensitive characteristic peaks in the spectral curve of liquor.
\item Our proposed methods achieve more fantastic performance with insufficient training samples (such as 25\% or 16\%). Although the 1-D-Deep-CNN \cite{sang2022one} achieves outstanding performance with sufficient training samples, when the training ratio decreases, its performance drops more obviously and is instability. The performance and stability of our proposed $\text{Spectrum-BERT}_{\text{OMCM}}$ are both well maintained when the number of training samples is greatly reduced. This proves that our proposed "\normalem{\emph{pre-training \& fine-tuning}}" paradigm enables the model to extract potentially informative features of the spectral curve for subsequent specific downstream tasks, in the pre-training stage in an unsupervised manner based on unlabeled data.
\item The performance of four Spectrum-BERT variants verifies our intuition in section \ref{sec3.3}: $\text{Spectrum-BERT}_{\text{OMCM}}$ achieves the best performance, $\text{Spectrum-BERT}_{\text{All}}$ is second, and $\text{Spectrum-BERT}_{\text{Null}}$ is slightly better than $\text{Spectrum-BERT}_{\text{CLS}}$. Although BERT achieves great performance by leveraging [CLS] for classification in the field of NLP, our experimental results show that this strategy is not applicable in the field of spectral curve classification. The comparison of $\text{Spectrum-BERT}_{\text{CLS}}$ and $\text{Spectrum-BERT}_{\text{All}}$ shows that it is reasonable to leverage the entire feature representation sequence for the classification of the spectral curve. On the other hand, the performance of $\text{Spectrum-BERT}_{\text{Null}}$ is lower than $\text{Spectrum-BERT}_{\text{All}}$, which indicates that it is not a great choice to leverage randomly combined curve pairs as input but not use the training task to differentiate them. The optimal performance of $\text{Spectrum-BERT}_{\text{OMCM}}$ justifies our attempt to obtain latent features of a single spectral curve by abandoning the NCP task and allowing Spectrum-BERT to focus on the MCM task.
\item The model performance of $\text{Spectrum-BERT}_{\text{OMCM}}$, $\text{Spectrum-BERT}_{\text{All}}$, and 1-D-Deep-CNN shows an obvious decline trend when the training samples are reduced (e.g., 25\% or 16\%). This indicates that the NCP task is not suitable for our classification task, and there is indeed an optimization conflict between the NCP task and the MCM task in the pre-training stage. However, the natural advantage of the "\normalem{\emph{pre-training \& fine-tuning}}" paradigm makes Spectrum-BERT still achieve relatively stable performance in the case of small training samples (e.g., 25\% or 16\%). Compared with 1-D-deep-CNN, which requires a large number of labeled samples for training, our model is more lightweight and efficient.

\end{enumerate}

\subsection{Performance Exploration on Imbalanced Dataset}\label{sec4.5}

\begin{figure}
	\centering
	\includegraphics[width=2.5in,trim=0 20 0 20]{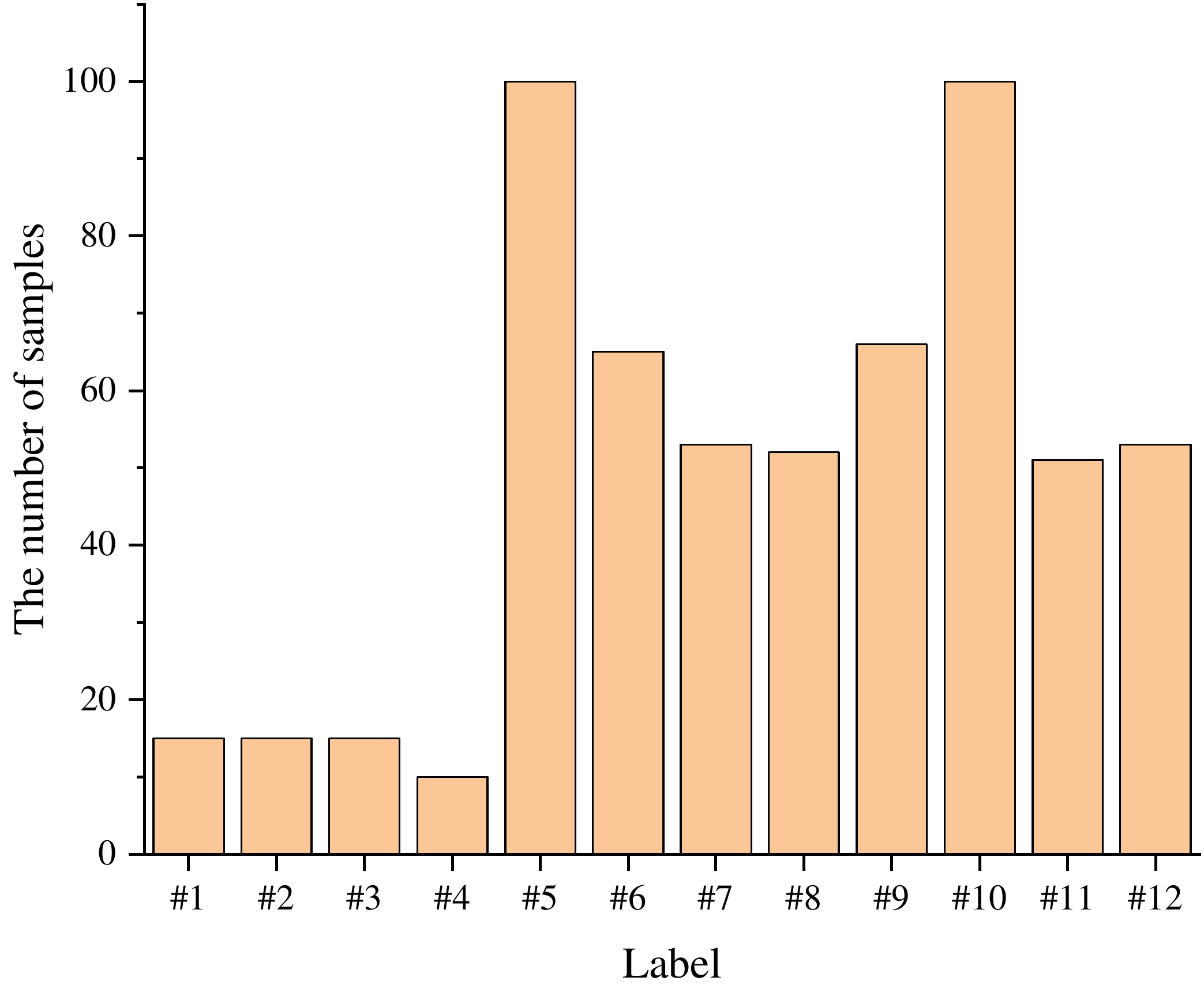}
	\caption{The sample label distribution of the constructed imbalanced dataset.}
	\label{FIG:7}
	\vspace{-1em}
\end{figure}

In most application scenarios, it is difficult for us to obtain a dataset with a relatively balanced distribution of sample labels. Unbalanced datasets are very common in the training phase of deep learning models. The performance of a model on an imbalanced dataset will directly determine its applicability in real-world scenarios. To test the performance of Spectrum-BERT on an imbalanced dataset, we construct an imbalanced dataset by randomly removing samples from the training samples, and the sample label distribution is shown in Fig.\ref{FIG:7}.

\begin{table}[h]
  \setlength{\abovecaptionskip}{-0.2cm} 
  \caption{The classification performance of Spectrum-BERTOMCM and baselines on imbalanced datasets(\%). Bold for “the best”.}
  \label{tab2}
  \setlength\tabcolsep{2.2pt}
  \renewcommand{\arraystretch}{1.3}
  \scriptsize
  \center
  \begin{tabular}{@{}cccccc|cccc@{}}
  \toprule
  Methods		& Precision
   & Recall		& weighted-F1  	\\ \midrule 

   \multirow{1}{*}{RF} 	 	 & $80.38\pm{17.97}$			& $78.42\pm{20.29}$  & $78.06\pm{22.26}$		 	
   \\ 

   \multirow{1}{*}{SVM}     & $86.12\pm{16.58}$		& $90.22\pm{0.12}$  & $86.41\pm{0.60}$		  
 	 \\
  
  \multirow{1}{*}{1-D-Deep-CNN} & $87.00\pm{101.65}$		& $90.29\pm{58.10}$   & $87.54\pm{96.50}$		  
	\\ 

   \multirow{1}{*}{BP}  & $91.37\pm{73.41}$		& $94.68\pm{35.94}$  & $92.69\pm{61.04}$		  
 	\\

  \midrule
   \multirow{1}{*}{$\text{Spectrum-BERT}_{\text{OMCM}}$}         
 & \bm{$96.54\pm{8.80}$} 	& \bm{$97.65\pm{3.32}$}   & \bm{$96.91\pm{6.29}$}	  
     \\ 
  
  \midrule
  
  \end{tabular}
  \end{table}

We train the baselines and $\text{Spectrum-BERT}_{\text{OMCM}}$ on the constructed imbalanced dataset, respectively, from the experimental results in Table \ref{tab2}, we can draw the following conclusions:

\begin{enumerate}[(1)]

\item $\text{Spectrum-BERT}_{\text{OMCM}}$ outperforms all baseline methods on imbalanced datasets by a significant margin. When the sample with different labels used in the fine-tuning stage are unevenly distributed, $\text{Spectrum-BERT}_{\text{OMCM}}$ can still fully capture the feature information of different samples, so that the model maintains high classification performance and good stability. This is important for deploying the model to real application scenarios.
\item 1-D-Deep-CNN, which is very competitive on balanced datasets, shows a significant performance decline on imbalanced datasets (Precision drops from 99.89\% to 87\%). This shows that the Max-Pooling operation used in 1-D-Deep-CNN makes the model lose the position sensitivity of the characteristic peaks of the spectrum curve. The loss of position sensitivity leads to 1-D-Deep-CNN can only judge the similarity of characteristic peaks but not the position of characteristic peaks. When the sample distribution is relatively uniform, this performance shortcoming can be compensated by utilizing the powerful representation ability of the complex model structure (as shown in Table \ref{tab1}). However, when the sample distribution is unbalanced (as shown in Fig.\ref{FIG:7}), 1-D-Deep-CNN cannot learn the feature representation of different substances through limited samples, resulting in large performance degradation and fluctuation.
\item The BP network, which does not perform well in the balanced dataset, outperforms other baseline models on the unbalanced dataset and achieves relatively excellent classification accuracy, but the results fluctuate greatly and the practical application value is limited. As a simple machine learning classification model, SVM achieves stable performance, but the classification accuracy of the model limits the application of SVM.

\end{enumerate}

\subsection{Hyperparameter Studies}\label{sec4.7}

\begin{itemize}

\item $token\_size$
	
\end{itemize}

In this section, we conduct exploratory experiments on the impact of different $token\_size$ on the classification performance of $\text{Spectrum-BERT}_{\text{OMCM}}$. The experimental results are shown in Table \ref{tab4}. After setting the $token\_size$ of the spectrum curve to 50 (smaller) or 200 (larger), the performance drops and becomes unstable compared to that is 100. We believe that when the block size is reduced to 50 points, the spectral curve is sliced into too small blocks. This may cause the characteristic peaks in the spectrum to be fragmented, which is unfavorable for Spectrum-BERT to capture informative features of the spectrum curve. On the other hand, when the block size is increased to 200 points, each block contains too much spectral curve characteristic information. This may lead to feature loss and feature omission when Spectrum-BERT captures features for each block. This proves that different $token\_size$ of Spectrum-BERT have a great influence on the model classification performance, and should be kept as small as possible without splitting spectral characteristic peaks. Our future improvement direction is how to make the model adaptively determine a more reasonable $token\_size$.

\begin{table}[h]
  \setlength{\abovecaptionskip}{-0.2cm} 
  \caption{The performance of $\text{Spectrum-BERT}_{\text{OMCM}}$ with different $token\_size$ (\%). Bold for “the best”.}
  \label{tab4}
  \setlength\tabcolsep{1pt}
  \renewcommand{\arraystretch}{1.3}
  \scriptsize
  \center
  \begin{tabular}{@{}cccccc|cccc@{}}
  \toprule
  	Method		& $token\_size$	& Precision
   	& Recall		& weighted-F1  	\\ \midrule 

   \multirow{3}{*}{$\text{Spectrum-BERT}_{\text{OMCM}}$} 	 	 & 50		
   	& $97.33\pm{6.44}$  & $97.11\pm{6.33}$	& 	$97.07\pm{6.34}$
   \\ 

    & 100			& \bm{$98.76\pm{0.25}$}  &  \bm{$98.62\pm{0.27}$}	& \bm{$98.61\pm{0.27}$}	  
 	 \\
  
    & 200			& $97.00\pm{15.46}$  & $97.16\pm{8.46}$	& 	$96.82\pm{14.02}$	  
 	\\
  
    \midrule
  
  \vspace{-0.7cm}
  \end{tabular}
  \end{table}

Additionally, we test the sensitivity of the classification performance of BASE sized $\text{Spectrum-BERT}_{\text{OMCM}}$ when setting different hyperparameter values.

\begin{figure}
	\centering
	\includegraphics[width=3.5in,trim=0 0 0 0]{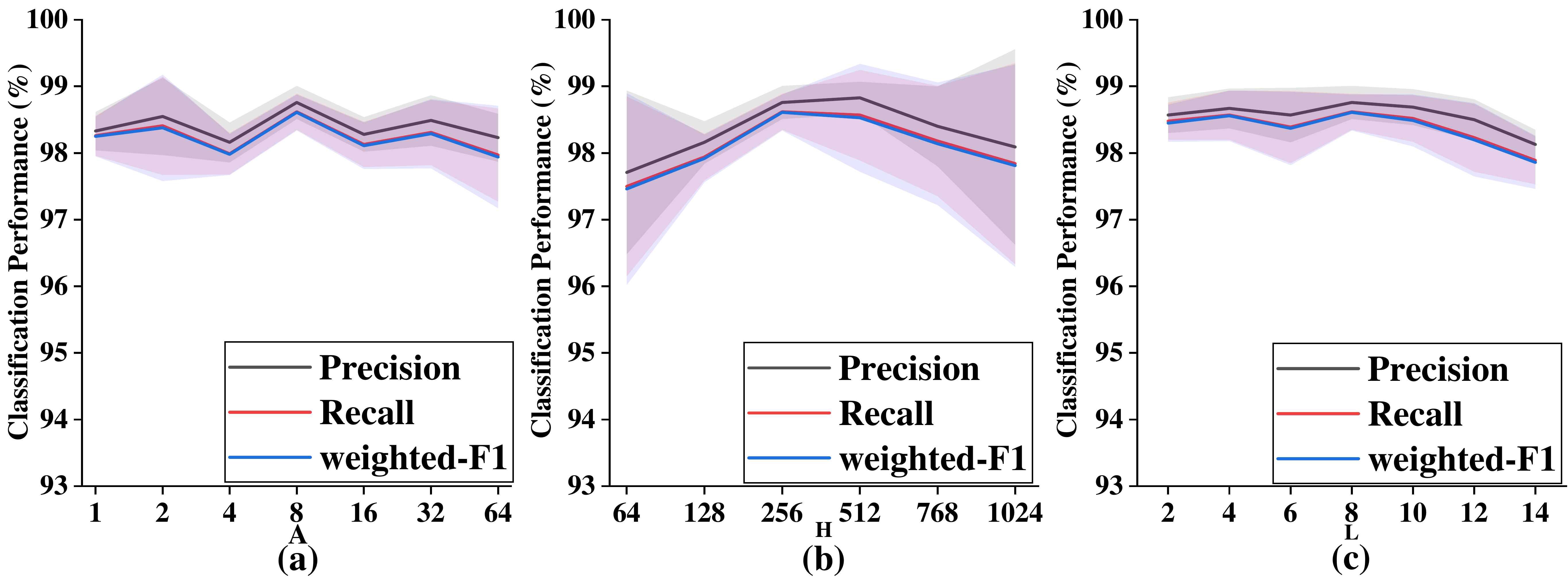}
	\caption{The performance of $\text{Spectrum-BERT}_{\text{OMCM}}$ with different values of hyperparameters. The curve is the average performance, and the error band shows the performance stability. We set the hyperparameters based on the results of the hyperparameter grid search experiments.}
	\label{FIG:8}
		\vspace{-1em}
\end{figure}

\begin{itemize}

\item $A$: The number of self-attention heads.
	
\end{itemize}

First, we fix the hidden layer size $H = 256$, and the number of layers $L = 8$ to explore the effect of different self-attention heads $A$ on the classification performance of $\text{Spectrum-BERT}_{\text{OMCM}}$. The experimental results are shown in Fig.\ref{FIG:8}(a). We can see that when $A$ increases, the model classification performance increases slightly but is unstable; the model classification performance begins to decline when $A$ increases to 16. This indicates that too few or too many self-attention heads increase the difficulty of capturing spectral curve features through training, which is not conducive to improving model classification performance.

\begin{itemize}

\item $H$: Hidden layer size.
	
\end{itemize}

Second, we fix the number of self-attention heads $A = 8$ and the number of layers $L = 8$ to explore the effect of different hidden layer sizes $H$ on the classification performance of $\text{Spectrum-BERT}_{\text{OMCM}}$. We increase the $H$ from 64 to 1024, and the classification performance is shown in Fig.\ref{FIG:8}(b). The overall trend of the performance curve is similar to that of the self-attention head number experiment, reaching a stable peak value when the $H$ is 256 (note that the model classification performance is slightly higher when $H=512$, but the fluctuation is greater than $H=256$), the performance curve shows a clear downward trend and is unstable on both sides of $H=256$. The experimental results show that the too-large hidden layer size not only increases the computational budgets of the model (16 times when $H=1024$ than that of $H=256$) but also introduces unnecessary noise, which is not conducive to the improvement of model performance. When the size of the hidden layer is too small, the classification performance of the model also declines significantly. We believe that the smaller hidden layer size is not enough to fully represent the captured spectral curve features, and lead to feature missing.

\begin{itemize}

\item $L$: Number of model layers.	

\end{itemize}

Last, we fix the number of self-attention heads $A = 8$, and hidden layer sizes $H=256$ to explore the effect of different model layers $L$ on the classification performance of $\text{Spectrum-BERT}_{\text{OMCM}}$. The experimental result (as shown in Fig.\ref{FIG:8}(c)) shows that different from the experiments on the other two hyperparameters($A$ and $H$), the number of model layers $L$ has little effect on the classification performance of $\text{Spectrum-BERT}_{\text{OMCM}}$, and the overall performance curve is stable.

\section{Conclusion}

Facing the Few-shot Learning problem in spectral curve classification of Chinese Liquors, this paper proposes the "\normalem{\emph{pre-training \& fine-tuning}}" paradigm for the first time in the field and builds the Spectrum-BERT model. We innovatively partition the curve into multiple blocks and obtain the embeddings of different blocks as the feature input to retain the model's sensitivity to the characteristic peak position and local information of the spectral curve. The two well-designed pre-training tasks, NCP and MCM, enable the model effectively utilize unlabeled samples to capture the potential knowledge of spectral data, breaking the restrictions of the insufficient labeled samples, and improving the applicability and performance of the model in practical scenarios. As we expected, Spectrum-BERT significantly outperforms the baselines in multiple metrics, and this advantage is more significant on the unbalanced dataset; in the hyperparameter studies, we also analyze the impact of different important parameters on the performances under different settings, providing references for subsequent related research.

In the future, we will improve the block partition strategy of the spectrum curve, so that the model can adaptively determine a more reasonable $token\_size$ and reduce the impact of human factors on the model. In addition, we notice that in the hyperparameter studies in section \ref{sec4.7}, when the hidden layer size $H$ is 64 or 1024, the performance of Spectrum-BERT fluctuates obviously, which is need to be optimized in the follow-up work. At the same time, we plan to leverage latent knowledge distillation \cite{wang2021knowledge} to explore how to better utilize the local feature representation of spectral curves obtained by Spectrum-BERT, so that Spectrum-BERT can perform more efficient computations with limited computing resources \cite{han2021dynamic}.


\bibliographystyle{IEEEtran}
\bibliography{IEEEabrv,cas-refs}











\vspace{-3em}
\begin{IEEEbiography}[{\includegraphics[width=1in,height=1.25in,clip]{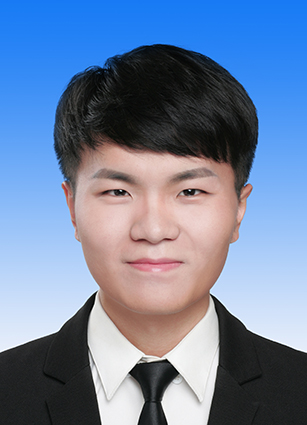}}]{Yansong Wang}
Yansong Wang is currently working toward a master's degree in the School of Computer Science and Technology at Harbin Institute of Technology. His research interests include machine learning, graph representation learning, recommender system, and spectral data analysis.
\end{IEEEbiography}
\vspace{-4em}

\begin{IEEEbiography}[{\includegraphics[width=1in,height=1.25in,clip]{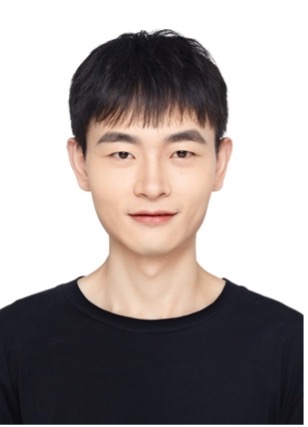}}]{Yundong Sun}
Yundong Sun is currently working toward the Ph.D. degree in the Department of Electronic Science and Technology at Harbin Institute of Technology. His research interests include machine learning, graph representation learning, recommender system, and spectral data analysis. He has published several articles in journals such as TKDE, IoTJ, Neurocomputing, etc.
\end{IEEEbiography}
\vspace{-3em}

\begin{IEEEbiography}[{\includegraphics[width=1in,height=1.25in,clip]{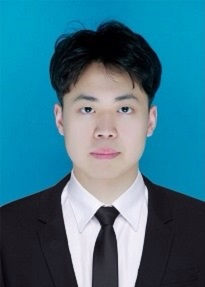}}]{Yansheng Fu}
Yansheng Fu is currently working toward a master's degree in the School of Computer Science and Technology at Harbin Institute of Technology. His research interests include multi-modal representation learning, machine learning, and graph neural network.
\end{IEEEbiography}
\vspace{-3em}

\begin{IEEEbiography}[{\includegraphics[width=1in,height=1.25in,clip,keepaspectratio]{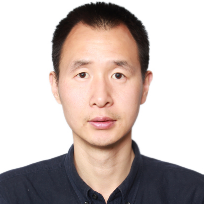}}]{Dongjie Zhu}
Dongjie Zhu received the Ph.D. degree in computer architecture from the Harbin Institute of Technology, in 2015. He is an associate professor in the School of Computer Science and Technology at Harbin Institute of Technology, Weihai. His research interests include machine learning and spectral data analysis. He has published more than 60 articles in several journals or conferences.
\end{IEEEbiography}
\vspace{-3em}
\begin{IEEEbiography}[{\includegraphics[width=1in,height=1.25in,clip]{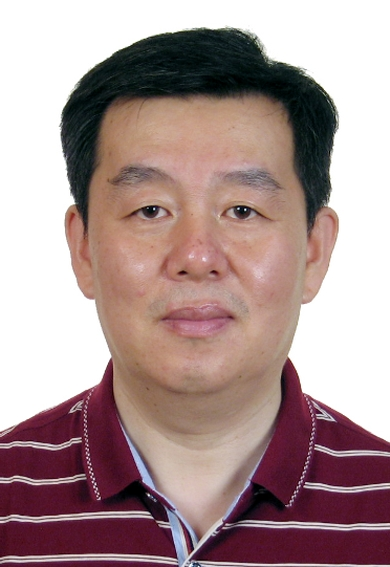}}]{Zhaoshuo Tian}
Zhaoshuo Tian is a professor in the Department of Electronic Science and Technology at Harbin Institute of Technology. His research interests include laser technology and marine laser detection technology. He is a member of IEEE.
\end{IEEEbiography}

\vfill

\end{document}